**Manuscript Title: Artificial Intelligence, Minds, and Brains**

**Article Type: Review Article**

**No Competing/Conflicts of Interest**

Corresponding Author:
Dr. Jay Seitz
Author e-mail:
drjseitz@mac.com
Neurocognitive Therapeutics, LLC
United States

**Abstract**

In the last year or so (and going back many decades) there has been extensive claims by major computational scientists, engineers, and others that AGI (artificial general intelligence) is 5 or 10 years away, but without a scintilla of scientific evidence, for a broad body of these claims: Computers will become conscious, have a "theory of mind," think and reason, will become more intelligent than humans, and so on. But the claims are science fiction, not science.

This article reviews evidence for the following three (3) propositions using an extensive body of scientific research and related sources from the cognitive and neurosciences; evolutionary evidence; linguistics; data science; comparative psychology; self-driving cars, and robotics; and the learning sciences. (1) Do computing machines (LLMs) think or reason? (2) Are computing machines (LLMs) sentient or conscious? (3) Do computing machines (LLMs) have a theory of mind?

_______________

Corresponding author: drjseitz@mac.com

# Table of Contents

4. Do computing machines have a "theory of mind?"

## 1. Preface

## A. Problems with the Use of AI Language

Throughout this article I examine the use of language when referring to the putative cognitive and affective capabilities of computing machines. In cases where such references are made with regard to computing machines, large language models (LLMs) and their computing substrates or any other artifactual information-processing system, use of such words and linguistic expressions are highlighted (" ") as it is implicit in their use that they are strictly metaphorical expressions without contravening evidence.

So, for example, 'thinking' and 'reasoning' include terms like 'problem-solving', 'deliberating', 'deciding', 'judging', 'planning', 'understanding', 'interpreting', 'recognizing', 'remembering', and 'recalling'. As well as cognate terms such as 'semantic reasoning', 'symbolic thought' or 'symbolic manipulations', 'sapience' or 'wisdom', as well as references to 'mental concepts', 'mental' and 'cognitive' 'processes' or 'cognitive skills' or 'abilities'. Related terms include such terms as 'learning' and 'knowledge' (acquisition).

There are serious problems, however, in the use of these terms as they implicitly impute causal powers to computing machines. So, to say that an LLM and its associated computing hardware is "thinking" or "reasoning" or engaging in "semantic reasoning" (tautological) or "recognizes," "recalls," or "remembers" when, in fact, it is merely accessing information in its data stores. Or an "inference engine" that applies logical rules (i.e., algorithms) to data to come up with instances of novel facts may be useful in drug discovery or myriad other areas. But the fact that it is "predicting"—only in the sense that the clock will strike twelve at noon—a possible new fit of molecules that may have some medical use, it is not "deciding" anything as that will be determined by the molecular biologists using the computing machine. Or, to put it another way, the "tools" aren't thinking, deciding or recalling but the scientists or programmers that use the tools. That's why the tools are useful. We think with them (see discussion below).

In a similar sense, an "expert system" comprised of an "inference engine" with access to a "knowledge base" (i.e., structured data organized and coded by human programmers) is not "thinking" or "reasoning" but applying algorithmic rules to data to derive new facts from established occurrences. The same applies to newer systems that use machine "learning" and data mining. But these large systems are stymied by computational issues as they are overcome by massive collections of rules.

## B. Use of the Terms 'Sentience' and 'Consciousness'

With regard to sentience and consciousness, sentience is the capacity to experience sensations—sensing or recognizing internal or external stimuli—as well as the ability to perceive emotions in oneself or others. It is the "simplest or most primitive form of cognition, consisting of a conscious awareness of stimuli without association or interpretation" (APA Dictionary of Psychology: https://dictionary.apa.org/sentience).

Consciousness refers to "an organism's awareness of something either internal or external to itself" (APA Dictionary of Psychology: https://dictionary.apa.org/consciousness). It consists in the awareness of the external environment (i.e., primary consciousness), but it may also include awareness of one's own internal states or self-awareness (i.e., secondary consciousness).

As I will argue in this article, however, current computing machines are not sentient or conscious nor do they think or reason or have a theory of mind. The question of whether they will become sentient or conscious, think or reason, or have a theory of mind in the future is pure speculation and otherwise meaningless in the present context. Alan Turing suggested a similar response to the question of whether "machines can think?" The answer might "be sought in a statistical survey such as a Gallup poll," which, he noted at the time, was absurd (Turing, 1950).

## C. Mind of a LLMentalist

A "mentalist" is a performing artist who appears to manifest highly developed intuitive or cognitive abilities. But they are not psychics, *per se*, and many eschew supernatural forces that psychics believe. Rather, they are adept purveyors of human psychology and produce—on "stage"—experiences for the imagination that expand reality—or are thought to do so—using influence and suggestion.

Baldur Bjarnason (Bjarnason, 2023), a technical services manager in Iceland, has written on the *intelligence illusion* when engaging with Large Language Models (LLMs). His central point is that LLMs capitalize on the same method of creating psychological illusion as does a real psychic or mentalist. A psychic, of course, is a person who claims to use extrasensory perception to identify facts about a person that are hidden from normal observation and the psychic plays on humans' natural susceptibility to self-delusion. But there is no proof of the existence of psychic abilities and the field is considered a pseudoscience (Druckman, D. & Swetz, J. A., 1988). Nonetheless, it is similar to the Barnum or Forer effect—the "fallacy of personal validation"—where vague and general descriptions of someone's personality, which could apply to just about anyone, are endorsed by the very same individuals given those descriptions (APA Dictionary of Psychology: https://dictionary.apa.org/barnum-effect).

Moreover, playing on humans' susceptibility to self-delusion is an element of psychological persuasion, a well-researched field. Psychological persuasion accomplishes its goals by targeting two inherent features of the human cognitive system: the way people think, and the way people feel or emote. With regard to the former, individuals typically engage in stereotypical ways of thinking and acting that have not been subject to critical thought, which forms the basis of prejudice and stereotypes. When you are able to get a person to act from their ingrained beliefs and attitudes you have effectively captured their minds because you have short-circuited critical thought. In a nutshell, critical thought entails examining the basis of one's thoughts and how one arrives at one's decisions and for what reasons (Seitz, 2018). And in the case of the hyperbole surrounding "artificial intelligence," the massive amount of monies involved to power large server farms (now reported to be the size of India's power needs)—among other things—the representation of AI in popular media such as science fiction films, and the rush to improve and insert AI into every aspect of life. Take, for instance, the recent claim that "there is no question that AI will eventually reach and surpass human intelligence in all domains" tweeted by an ACM Turing Award Laureate (4.15.2024). Essentially, a meaningless statement, entirely speculative, and ironic since we don't even fully comprehend intelligence in humans or other animals, it's brain and bodily basis, and how it is shaped by the social and cultural environment. Or, to put it another way, to what domains are we actually referring?

People's expectations, however, often play an even more powerful role than the content of a spoken or tweeted message or display of a visual image. This is so because people's prior beliefs and attitudes form powerful expectations of what they anticipate might *actually* happen. This occurs because people discount or ignore the influence of factors that do have tremendous influence including the powerful role of appeals to the emotions, industry authority figures, the credibility of sources (e.g., a corporate executive of a large technology company); and the influence of the herd instinct (e.g., Internet stock market bubble in the late 1990s).

In the case of LLMs, or what Bjarnason calls the "LLMentalist Effect," the audience tends to be self-selected (e.g., early adopters, tech enthusiasts, genuine believers in AGI), the chat environment sets up expectations and mood, a prompt is presented to the chatbot and users who engage in conversation become more invested, the answers are generic but the chatbot delivers a statistically plausible response and, as a result, the "mark" gets even more invested and becomes, in essence, a victim of a "confidence trick." Of the latter, the mark's naivete and credulity are now a clear target and become part of a "subjective validation loop" in which the more they engage they are more convinced of the chatbot's intelligence. Finally, the "mark" is left with the feeling that the chatbot is self-aware and capable of reasoning. In some ways, a similar process is applicable to the cryptocurrency enterprise.

Given the limits on human reasoning, particularly errors in judgment and decision-making including the universality of cognitive biases, the LLMentalist Effect is understandable. Among other things, these cognitive biases include confirmation and projection biases, that is, recalling information in a way that confirms one's prior beliefs or projecting current beliefs onto a future event, overconfidence effects, bounded rationality or the limits on human reasoning and irrationality, and agent detection or presuming the purposeful intervention of an intelligent agent in situations where none exists. Moreover, just about anyone is susceptible to the effects of these so-called confidence tricks because they are major features of human cognition.

## 2. Do Computing Machines Think or Reason?

## A. What are Thinking and Reasoning?

So, I am making three (3) claims in this paper.

1. Computing machines (LLMs) do not think or reason.

2. Computing machines (LLMs) are not sentient nor conscious.

3. Computing machines (LLMs) do not possess a theory of mind.

First of all, when we attribute thinking or reasoning to a person or thing, what do we actually mean?

Thinking and reasoning imply conscious, cognitive processes (see p. 23 for the discussion of empirical findings that consciousness is necessary for logical reasoning in humans). That is, if I'm thinking about something, I'm aware about what I am thinking about—not necessarily the underlying mental processes that I use to think with—but rather the results of those mental processes.

Indeed, I can even think about my own thought processes and how they might be related to each other in some way, known as metacognition.

Therefore, to say that someone or something is thinking or evincing thought, is to say that they are aware that they are thinking and that their thinking is from a particular viewpoint, that is, from the perspective of the someone or something doing the thinking. Hence, *it must follow that all thought has a subjective aspect,* that it is from some unique perspective or point of view.

However, an individual could be experiencing a delusional state or delirium (a neurocognitive disorder in DSM-5 recognized as an acute confusional state) or an altered state of consciousness produced by a psychoactive drug or some other external experience (e.g., a traumatic event) or pathological brain event, but these various states are signs of underlying pathology, which would affect ongoing processes of thought. So we will exclude them here for consideration.

Nevertheless, the cognitive and neurosciences clearly distinguish between conscious cognition, what I am thinking about right now and currently in working memory, and nonconscious cognition or what is potentially accessible in long-term memory such as reporting my city of birth (semantic memory) or what I ate for dinner last night (episodic memory). In this sense, I refer to nonconscious cognition as an assemblage of "underlying mental or cognitive processes" that support conscious cognition (see p. 25 for a discussion of the related concept of latent "computational anatomy" in humans and other animals).

Indeed, these underlying "mental or cognitive processes" can also be construed as various kinds of "computations." So, for instance, I can use a HP hand calculator to solve complex algebraic equations using rules or "algorithms" that operate over formal logical principles. But is the HP hand calculator engaging in "thought" (conscious or otherwise) as a few buttons are pressed?

> "An algorithm is an explicit, step-by-step procedure for answering some question or solving some problem. An algorithm provides *routine mechanical instructions dictating how to proceed at each step.*" – Stanford Encyclopedia of Philosophy (SEP, 2020).

But it does not follow that just because a computer can "solve" a problem using algorithm(s), that it is engaging in thought, conscious or nonconscious. Yet humans can consciously intervene in underlying nonconscious physiological processes, that is, in the "opposite direction," using conscious thought to effect nonconscious processes, using bio- and neurofeedback (see pp. 10-12 below for a discussion of relevant empirical findings).

Indeed, according to the *classical computational theory of mind* (CCTM), the mind/brain is a computational system whose "symbolic manipulations" or "mental processes" are similar to a Turing machine (SEP, 2020).

Nevertheless, a Turing machine is, at best, an extremely simplistic model of how the mind/brain actually works. The very first Turing machine manipulated symbols on a strip of tape according to a table of rules. And these rules are referred to as algorithms. But modern computers have the essential computational features of Turing machines, that is, the ability to compute the output of a mathematical function given a specific input.

Alan Turing proposed that if a human evaluator in a staged conversation could not tell a machine from a human, then the computer had demonstrated human-like intelligence (Turing, 1950). But the ability of a Turing machine to convince someone to mistake it for a human—an "imitation game"—completely misses the point.

1. For one, it assumes that thought is a form of computation whereas thought (conscious or unconscious) is the result of that computation (e.g., stringing syntactic expressions together, articulating movements in the larynx, connecting two brain areas in unison) but not thought itself. So, the computer is not literally engaging in conversation, it's merely simulating, *mutatis mutandis*, what appears to a human to be conversation.
2. If I hook up Wikipedia to a talking e-reader does that mean that the encyclopedia is engaging in thought because it actually speaks to me?
3. If I engage in conversation with an AI program like Perplexity (or Anthropic's Claude or Google's Gemini or Microsoft's Copilot) and it responds with an answer to my query, does that mean it has an inherent ontology or worldview about what the world is actually like?
4. If a human believes that a computer is human, what are the transference-countertransference relations? which is also an utterly absurd question.

Let's look at bio- and neurofeedback as a case in point.

## A1. Biofeedback and Neurofeedback

In biofeedback and neurofeedback many brain and bodily functions occur at a level of nonconscious awareness. But a conscious human can potentially intervene in these nonconscious processes and alter their course.

1. So, an individual can acquire conscious awareness of many underlying nonconscious physiological functions using various instruments (e.g., EKG, EEG, EMG: heart, brain and musculature, respectively) that provide feedback on ongoing activity of brain and bodily organs with the goal of being able to manipulate them mindfully. To manipulate them mindfully—an introspective or meditative ability—is to acquire a cognitive/symbolic skill to intervene in these underlying nonconscious physiological functions. That is, effect them from a level of conscious awareness even though the conditioning processes themselves may be unavailable or nonconscious to the individual.

   For example, in one study (Mehler et al., 2018), researchers trained brain areas involved in affective processing using functional magnetic resonance imaging neurofeedback (fMRI-NF) to reduce patient's depressive symptoms. This kind of neurofeedback enables individuals to develop self-regulatory strategies that impact brain networks associated with mental imagery. And the patient's enlistment of mental imagery can increase their cognitive flexibility and facilitate kindling of positive mental experiences. As well as further enhance the patient's monitoring and feedback from associated brain areas that reflect their own underlying neural activation patterns.

   In another study, neurofeedback training was employed to teach athletes to regulate the activation patterns of targeted regions of the cerebral cortex (Yiang et al., 2018). Athletes' cortical electroencephalograms (EEG) were continuously monitored, and feedback of their brain activity was transformed into visual or auditory form. With extensive training, athletes learned that specific mind/brain states correlate with certain EEG signals allowing them to voluntarily control areas of the cortex related to emotion, cognition, and behavior by consciously increasing or decreasing the power of targeted EEG frequency bands.

But there is no evidence (excepting neurosymbolic approaches) that any existing large model artifactual information processing system operates on formal "symbols"—the symbols themselves are mere placeholders—and only simulate actual mental states. But they are not the "states" or the results of those states themselves. That is because there is no layer above the hardware and its operational software that supervenes on the underlying computational processes (cf. above for discussion of empirical findings on the use of bio- and neurofeedback that operate on underlying nonconscious processes).

## A2. *Large Language Models (LLMs)*

Large language models (LLMs) acquire the ability to generate language by tirelessly accumulating *statistical relationships among text documents*. And they are trained on vast amounts of text using recurrent networks that are capable of both feed-forward and feedback mechanisms (i.e., backward propagation). These "artificial networks" are bidirectional insofar as the output of nodes in the computer architecture feedback on subsequent input from those very same nodes. That way, inputted text can predict the next word or sentence or statistically link to other elements in the overall text. But multimodal systems can also be trained in a similar fashion on images or even video or motion.

In a comparable manner, natural language text can be structured so as to describe a task that the computer should perform using what is known as "prompt-engineering." In this way, a query about the definition of 'exaptation' or a statement about the best tabouleh salad, and many, many other use cases, can be performed.

However, generative LLMs often produce output that is not justified based on their training data, a phenomenon known as a "hallucination." But the notion of a "hallucination" is derived from human phenomena where there is perception of something even though it doesn't actually exist. Formally defined as a "false sensory perception that has a compelling sense of reality despite the absence of an external stimulus" (APA Dictionary of Psychology: https://dictionary.apa.org/hallucination).

Moreover, critics of LLMs believe that they are merely recombining and remixing text and possess serious deficits in prediction and reasoning skills, planning, and learning—thus referred to as "stochastic parrots"—so that they kind of sound like what a real person might have written. Edward Tian, a recent Princeton graduate, has developed an "AI detector" that differentiates between human-produced and AI-produced text (www.GPTzero.me) and probabilistically is surprisingly accurate.

"A system like ChatGPT [GPT-3] doesn't create, it imitates. When you send it a request to write a Biblical verse about removing a sandwich from a VCR, it doesn't form an original idea about this conundrum; it instead copies, manipulates, and pastes together text that already exists, originally written by human intelligences, to produce something that *sounds* like how a real person would talk about these topics…"

…But, once we've taken the time to open up the black box and poke around the springs and gears found inside, we discover that programs like ChatGPT don't represent an alien intelligence with which we must now learn to coexist; instead, they turn out to run on the well-worn digital logic of pattern-matching, pushed to a radically larger scale."

– Cal Newport, Associate Professor of Computer Science, Georgetown University.

Nevertheless, there is no "ontology" in the computer or its operation, and it is not engaging in symbolic thought because there are no "symbols" in the computer to process or any other mechanism within its hardware to carry that out. Ontology, of course, is inquiry about the actual state of the world, the types of entities that exist in it, how they may be grouped together, and in what sense they might be related to one another.

Let's see why this is so.

## A3. Self-Driving Cars

### Case Study #1

Self-driving cars pose an interesting problem. With an array of sensors around the exterior of the car and some limited computing in the car along with GPS coordinate feeds from the sky, they set out to drive themselves down the busy streets and highways of North America and beyond under just about any and all environmental and weather conditions.

Moreover, driving modes in self-driving cars range from basic driver assistance to partial, conditional, high, and full automation of the car where the driver assistance system controls the vehicle under all driving conditions. Partial automation is currently found in about 50% of US cars and provides driver assistance for lane change correction (i.e., crossing the lane line), crash prevention forcing the driver to apply the brakes, adaptive cruise control, monitoring the front and the rear of the car for pedestrians and vehicles, blind spot monitoring, and so on.

But the deployment of autonomous driving technology is encumbered by, among other things, "corner" or "edge" cases which the software cannot "grasp" or act upon. So, one argument is that such corner cases arise from what is called a "narrow AI" approach which grossly lacks much of the general knowledge base that any normal human makes use of while driving. That is to say, autonomous driving technology is completely bereft of any inherent ontology about what the world is actually like.

In a recent study (Da Lio et al., 2023), affordance competition in layered control architectures was used to create an artificial agent devoid of algorithms but, nonetheless, capable of adaptive behaviors in real-time. Affordance, of course, is the capability of a person or artificial system to perceive important features of the environment and engage in possible actions to take advantage of it.

"The difference between narrow and general intelligence becomes crucial when dealing with edge- and corner-cases. During unexpected situations and emergencies, the most appropriate reaction might come from a broad knowledge of objects in the world and intuitive cognition of the physics that rules their motion and interactions" (Da Lio et al., 2023).

### A4. Robotic Models

### Case Study #2

Let's take Google's new robotics model, RT-2 (New York Times, July 28, 2023; https://robotics-transformer2.github.io/ and https://deepmind.google/discover/blog/rt-2-new-model-translates-vision-and-language-into-action/). In reality, it's just a very primitive robot with optic sensors that can pick up an object and manipulate it—very clumsily to be sure—and quite crude compared to Boston Dynamics "embodied humanoids" (see below) and it also has "pattern detection abilities" to respond to human language. That is, it links a large language model (LLM) to the robot's physical manipulations and ties it to its optic sensors that can "recognize" (i.e., code) images, which are quite limited, too. But that is really the only new thing here in robotics, the LLM-physical manipulation link, as the study of "robotics" itself has been around since the third century B.C.E. in early descriptions of automata.

And calling something "semantic reasoning" or "chain of thought reasoning" or "multi-stage semantic reasoning" does not make it so. Notice the use of words (excerpted below). 'Semantic' implying it's acquiring meaning from its reasoning, which is tautological; that it 'learns' rather than just acquires more data, and that it "maps…observations to actions" rather than the machine acquires information to link robotic data to action data so as to "enjoy" the benefits—rather than achieve a successful outcome—of what the software program has acquired.

"We study how vision-language models trained on Internet-scale data can be incorporated directly into end-to-end robotic control to boost generalization and enable *emergent semantic reasoning*. Our goal is to enable a single end-to-end trained model to both learn to map robot observations to actions and enjoy the benefits of large-scale pretraining on language and vision-language data from the web."

And calling something "as figuring out what object to pick up," rather than generalizes to a hammer or energy drink, does not make it so.

…We further show that incorporating chain of thought reasoning allows RT-2 to perform *multi-stage semantic reasoning*, for example figuring out which object to pick up for use as an improvised hammer (a rock), or which type of drink is best suited for someone who is tired (an energy drink)" (Brohan et al., 2023, July 28).

But to be clear, Google's robot doesn't "understand" anything and it's smart only in the sense that a HP hand calculator is "smart." That is, it can make use of complex algorithms or rules that operate over formal logical principles. The actual underlying computations are carried out by integrated silicon circuits that utilize two subatomic electrical parameters: off and on. It's a binary representation of the world and the binary bits only stand for things at the computational level.

What are these computational levels?

### A5. Levels in a Computational System

David Marr (1982) proposed three, now four levels of analysis, in describing any "computational" system that converts raw sensory information into concepts and ideas. For instance, we can think of our biology as processing or transducing physical information in the environment to more useful electrochemical information processed by the mind-brain-body. This first level is the "physical" implementation of a sensory modality in a biological system. In vision, the processing of light by the human eye and the transmission of this biochemical information by way of the optic nerve and the optic pathways is assembled in the visual or striate cortex at the back of the brain. But what algorithms or rules does the brain use to create three-dimensional vision in its wet architecture? This second level is the algorithmic realization of vision, in this case the myriad processing areas of the visual cortex (e.g., color, line, form). And the third level is the computational level. How does the brain assemble shapes, colors, and diverse spatial perspectives—in Rudolf Arnheim's language (Arnheim, 1969), perceptual forms or "percepts"—to identify objects and things in the real world and imbue them with meaning? That object over there is an apple. That moving shape over here is a giraffe. Indeed, this fourth level is the way that culture and environment shape the process of seeing and sensing in other modalities. We call this latter process, learning.

But computing machines (LLMs) don't "learn" anything in this sense, but only acquire new information (i.e., data), "skills" or preferences of various sorts that alter the course of their subsequent operation. But in *H. sapiens*, many aspects of learning are primarily about symbols— semiosis, meaning, and representation. So, one might say that computing machines largely acquire information—besides large "data dumps"— through association and reinforcement including statistical associations (see below), whereas humans and other organisms learn by way of a plethora of different mechanisms and experiences (Gardner, 1993, 2022; Gottfredson, 1996; Grandin, 2009: Sternberg, 1988; Vygotsky, 1978).

So, for Howard Gardner (1993), it's not how smart we are as individuals, but *how we are smart* that's critical, implicitly arguing against Aristotle's *scale naturae*, that is, an ascending scale in human beings from least to most intelligent. And humans learn in myriad ways undergirded by eight major kinds of intellective faculties comprising numerous sub-faculties and subordinate abilities and skills.

These include logical-mathematical abilities and skills; linguistic facility; visuospatial abilities; emotional and social adeptness in understanding one's self and others; musical proficiencies including understanding and making use of melody, harmony, rhythm, and sonority; naturalistic skills such as the ability to make distinctions with regard to flora and fauna; and bodily-kinesthetic or instrumental use of tools of various kinds including one's own body (athletics, dance) as well as objects external to the body (e.g., a car or set of winter skies).

Each individual has a unique panoply of intellective abilities "tuned" in development by the social and cultural environment that operates on basic bodily, biological/genetic, and neurological systems.

On the other hand, Lev Vygotsky (1978) argues that social learning plays a critical role in human development. Intellective abilities are forged through use of language (i.e., verbal mediation) and diverse kinds of tools which mediate between individuals and their social and cultural environments. Through language and other symbolic activities, human psychological processes became semiotically mediated and thus human cognition is vastly different from other species and, implicitly, any extant computing machine. In essence, individuals use a broad range of cultural "tools" that function as mediators in thought, reasoning, and memory.

"Importantly, the processes of activity are *first* mastered with an adult and *later* become internalized at the mental level. As such, psychological functions are initially *interpsychic* as an activity between a child and an adult or a more knowledgeable peer and only later become *intrapsychic* as individual thinking of a child. Consequently, Vygotsky devoted much attention to the concept of "inner speech" as a special type of psychological activity and suggests that speech develops first in the social environment and later becomes internalized into mental processes" (Vasileva & Balyasnikova, 2019).

Robert Sternberg (1988) proposed three major areas of human intelligence comprising practical or experiential, analytic, and creative or synthetic abilities. And use of our intelligences in each area incorporates one's "mental self-management skills." The latter includes planning and monitoring problem-solving, acquiring new knowledge through sifting out relevant from irrelevant information—*separating the wheat from the chaff*—combining information into a more connected whole, and relating new information to previously acquired knowledge. But the use of our "mental self-management skills" is shaped by real-world environments as well as experience in dealing with novel kinds of experiences and unexpected situations along with their execution in diverse contexts.

The historical or "standard" view of intelligence (viz., general mental ability) is captured in the concept of 'g', which is a composite of factors made up of correlations on various cognitive tasks (e.g., verbal fluency, mathematical abilities, memory, spatial visualization, vocabulary, and general information). These may be further differentiated into fluid and crystallized intelligence, processing speed, and related sub-factors. This received or standard view is endorsed by a number of social scientists in the field and others but has been challenged by a broad range of views (above) since the 1980s. Nonetheless, a number of long-standing aptitude and achievement tests are partly based on it.

Notwithstanding, Dr. Temple Grandin documents insights she has gained from her personal experience of autism. She believes that there are three major kinds of thinking in humans: Photo-realistic visual thinking, pattern thinking, and word-fact thought.

Of the first kind, "I can see everything in my head and then draw it on paper," but such individuals are often poor in algebra but much better at geometry and trigonometry. Of the second kind, "pattern thinkers see patterns and relationships between numbers," but reading and written composition are poor. Of the third kind, such "individuals have a huge memory for verbal facts" but are poor at visual thinking such as drawing (Grandin, 2009).

What are "pattern thinkers?"

## A6. Pareidolia

Pattern detection abilities—technically "pareidolia"—are presumed to have evolved in land animals at least 520 million years ago during the Cambrian period, known as the "Cambrian explosion," in which there was rapid diversification of animal lifeforms as well as probably much earlier in the highly diverse biological forms that originated in the oceans (Seitz, 2019). Indeed, primary consciousness evolved during this period (Feinberg & Mallet, 2017) as land animals became conscious of their physical environment and set in motion its use as a prediction device. For more on the evolutionary origins of pareidolia, see Seitz (2019; https://lnkd.in/eZVhv28).

But pattern recognition is actually an ancient feature of plant-based immunity. In plants, pattern-triggered innate immunity (PTI) is triggered by microbial patterns via cell surface-localized pattern-recognition receptors (PRRs). Conceivably, originating as early as the Precambrian era, 850 million years ago (Yuan et al., 2021).

So, pareidolia is widespread in the animal and plant kingdoms and in humans includes recognition and recall memory for faces, music, natural (e.g., voice) and unnatural sounds, words, nonverbal communication (e.g., gait, use of the hands and body), a vast array of "images" (e.g., photographs, drawings, paintings, sculptures) and so on. "Image detection" in "computing machines" is a very recent development and highly circumscribed. But let's be clear, not a single image-bearing radiologist has been replaced by artificial "intelligence."

"We often compare AI algorithms to radiology experts based on the ability to identify a single disease or a small set of diseases. These assessments dramatically oversimplify what radiologists do. A comprehensive catalog of radiology diagnoses lists nearly 20,000 terms for disorders and imaging observations and over 50,000 causal relations. An AI algorithm that diagnoses common chest conditions at the level of a subspecialty thoracic radiologist is a major step forward…But human radiologists are also trained to detect uncommon diseases in the long tail of the distribution, including rheumatoid arthritis, sickle cell disease, and post-transplantation lymphoproliferative disorder. AI is impressive in identifying horses but is a long way from recognizing zebras" (Langlotz, 2019).

Let's turn to the critical importance of embodiment.

### A7. Embodiment

We do not simply inhabit our bodies; we literally use them to think with. And this occurs through our use of the perceptual system to sense and react to the physical world, through the use of our motor system which allows us to act on that physical world, as well as our mental and bodily interactions with the current culture and environment.

For example, Boston Dynamics' "embodied humanoids" are quite impressive in many ways (https://bostondynamics.com/). Unfortunately, the *computational theory of mind* fails to understand the importance of, and incorporate embodiment and its centrality, in the ability of humans and other animals to physically manipulate "tools" in their environment and directly interact with their world (Seitz, 2005a; https://lnkd.in/gUZ8Tehe).

One of their "humanoid robots," *Atlas*, has enhanced agility such as the ability to spin its body completely around, athletic prowess like jumping over objects or walking up steps, "dancing" in unison with other humanoid robots, the ability to engage objects with bimanual manipulation such as picking up and tossing large objects, and fairly advanced sensory-perceptual abilities. But while "androids" aesthetically resemble humans, a humanoid robot only roughly mirrors the shape of the human body typically possessing a head, arms, legs, and torso.

Similar agility is seen in the Tesla Optimus Gen 2 with fine-motor skills to enable picking up and handling an egg and with gross-motor abilities to walk at a fairly normal human gait, perform a squat with requisite balance, and cognate abilities: https://www.youtube.com/watch?v=sAW8xpgEKjM).

Or "Figure 02" just introduced by the startup Figure AI (https://www.figure.ai) which they refer to as a "general purpose humanoid."

"We've designed our world for the human form. Hands allow us to open doors and use tools; arms and legs allow us to move efficiently, climb stairs, lift boxes, and more. Figure 02 brings together the dexterity of the human form and cutting-edge AI to go beyond single-function robots and lend support across manufacturing, logistics, warehousing, and retail."

Nevertheless, their claim about "speech-to-speech reasoning" is misleading as it simply uses an LLM (GPT-4o) supplied by OpenAI for more polished prompt engineering.

Numenta (https://www.numenta.com/), the company spun off from Palm Computing (Jeff Hawkins), has laid out the minimum attributes for a machine "intelligence": Embodiment, multiple processing centers that operate collectively in building up a complex model of the world, continuous learning, and the importance of reference frames. Numenta's "reference frames" are the "subjective aspects of thought" (see discussion on p. 8 above). That is, thought can only be understood within the context of "some unique perspective or point of view" where one's knowledge of the world is always relative to one's distinct viewpoint.

But a LLM that passes the LSAT (Law School Admission Test) or the MCAT (Medical School Admission Test) is not thinking or have a point of view on anything. For instance, in one study, ChatGPT (GPT-3.5) performed at or above the median performance on the MCAT compared to 276,779 pre-medical students who took the test.

This is misleading, however, "since ChatGPT can only process text-based inquiries and encoding of visual items would introduce interpretive bias into our experiments, all sample test questions were manually screened, and questions necessitating visual analysis to answer them, such as graphs and diagrams, were removed" (Bommineni et al., 2023).

But to be sure, it's completely decontextualized information that's been dumped into its data stores and it uses stochastic processes/statistical relationships to ascertain what may be the answer to the next question.

Wikipedia partly solves this problem because it is written by human collaborators and vetted by a community of experts. So, why would anyone think that the LLM *itself* is intelligent or the machines that it runs on?

## A8. Statistical Learning and Reasoning

Thus, one consequential critique of machine "learning" systems is that they aren't participating in any actual thinking or thought process, but merely trading in statistical probabilities over time.

1. "Their deepest flaw is the absence of the most critical capacity of any intelligence: to say not only what is the case, what was the case, and what will be the case—that's description and prediction—but also what is not the case and what could and could not be the case. Those are the ingredients of explanation, the mark of true intelligence. Whereas humans are limited in the kinds of explanations we can rationally conjecture, machine learning systems can learn both that the earth is flat and that the earth is round. They trade merely in probabilities that change over time." (Chomsky, Roberts, & Watumull, 2023).

Yet, research has demonstrated that *logical reasoning* in humans is wholly dependent on *conscious cognitive processes* (DeWall et al., 2008). For example, nonconscious priming increases activation of logic-related concepts but not logical reasoning performance. Priming is a phenomenon in humans in which exposure to one stimulus may affect one's response to a different stimulus but without any conscious involvement. On the other hand, actually stimulating the conscious goal of reasoning leads to clear improvements in logical reasoning performance.

But computing machines aren't engaging in any sort of "logical reasoning" in this sense as they are merely instantiating in their hardware nonconscious software-mediated algorithms (i.e., *routine mechanical instructions dictating how to proceed at each step*). Because there is no layer above the silicon and its operational software that supervenes on the underlying computational processes (see the empirical research for the lack of evidence of logical reasoning in LLMs on p. 29 below).

Indeed, *statistical learning*—detecting meaningful regularities in seemingly random sequences—has been demonstrated for speech (words), non-speech sounds, static shapes and shape configurations, tones, and action sequences in human infants and adults (Seitz, 2019). Individuals are even able to track conditional and joint probability for dynamic human action by segmenting action units. The probability of an event given that another event has occurred is called a *conditional probability*. For example, the conditional probability of sneezing if you have an upper respiratory infection is quite high given the symptomatic relationship between the two. On the other hand, *joint probability* represents the frequency that any two elements co-occur and is usually expressed as the probability of event Y occurring at the same time as event X. So, if there is a rock or hip-hop concert (X) there is a high joint probability that young people (Y) will be in attendance.

On the other hand, dynamic actions are more difficult to segment because there is much more information to attend to and filter. Nonetheless, humans can segment and pick out these dynamic actions in order to understand enactive relationships, such as a curve ball thrown by a pitcher connecting with a bat swung by a baseball hitter.

Moreover, humans spontaneously segment dynamic action into hierarchically structured parts, such as a golf swing. What people segment has direct effects on what they tend to remember, what they learn from it, and the intentions behind it. They are not just learning *joint probability*—the co-occurrence of actions and their frequencies—but *conditional probability* or prediction of the relations between dynamic action elements, such as hitting the cue ball sharply towards its bottom edge in order to put backspin on the cue ball so that it returns to its original location after the shot is completed. That's *statistical reasoning*. And there is no extant evidence that LLMs are engaging in statistical reasoning, either, opposed to accruing statistical relationships, which in humans and other species is known as "statistical learning."

What about nonconscious processes that underlie cognitive faculties in biological systems?

### A9. Computational Anatomy

In animals—including humans—nonconscious mental or cognitive processes are embedded in their underlying biology. For instance, some animals such as dolphins and bats use bio-sonar ("echolocation")—the ability to maneuver and find prey in their environment—by using sound waves to locate obstacles and potential quarry. Dolphin sonar is composed of broadband cells in the brain that respond to a wide range of auditory frequencies and reaches well above the frequency range of normal human hearing—which tops out about 20,000 Hz—to 20,000 to 40,000 Hz. Indeed, sonic waves in the ocean convey sound energy at about 1376 meters per second, four times faster than sound energy in the air which ripples through the ether at about 344 meters per second.

Of the latter, bat sonar works in a similar range and certain moths that bats feed on have evolved the capacity to engage in evasive actions when bat sonar targets them. How does bat sonar work? When a bat approaches a target, let's say an edible moth, changes in the velocity of approach of the bat (the rate of change of her position) are represented in neural maps in the brain that change their firing pattern as the velocity changes. This is known as "computational anatomy." And it is nonconscious.

And there are biological compasses. Many species of avians make use of both sun and magnetic compasses to navigate their aerial or arboreal habitats. For instance, birds can make use of the position of the sun yoked to the time of day to fly in a certain direction. That is, they acquire a celestial reference point and use it to navigate.

Honeybees, on the other hand, rely on their sensitivity to ultraviolet light to navigate using the position of the sun. What appears to be involved is that the celestial pattern of polarization of sunlight is fundamentally the same as the pattern of polarization of the receptors in the bee's eyes. The bees simply match these two and the bee is off and flying to the appropriate food source or distant hive. The computational anatomy is embedded within the bee's optic system.

But in every case, these nonconscious biological programs are essential to survival and carried out automatically under the appropriate environmental conditions. There is no presumption of direct conscious intrusion.

And there are cognitive maps extensively studied in London taxicab drivers (Maguire et al., 2000). The driver must learn and memorize 320 standard routes through central London encompassing close to 25,000 streets within a six-mile radius of Charing Cross as well as all the major arteries throughout London. They must also acquire many points of interest including squares, clubs, hotels, restaurants, theatres, embassies, public and government buildings, houses of worship, police and railroad stations, cemeteries, crematoria, parks, sports centers, schools, historic buildings, as well as the order of theatres on Shaftesbury Avenue. Here many of these nonconscious maps are only accessed and placed in conscious thought when they are needed. A unique feature of human cognition and the primate brain as well as found in a number of other species.

Similarly, when an individual plays the piano, they don't have to think or reason about—or be conscious of—every key or set of keys they press, because nonconscious motor programs in the motor cortex and cognate areas carry out many of these functions nonconsciously ("without conscious thought"). The same can be said of innumerable aspects of speaking and many other cognitive and motor activities carried out by the brain in communion with the body.

## A10. Creative Thought

It was announced in 2022 that AI outperformed human subjects on a creativity test. Artificial "intelligence" (i.e., GPT-4), was found to match the top 1% of "human thinkers" on a standard creativity test (i.e., TTCT, Shimek, 2023).

The Torrance Tests of Creative Thinking (TTCT), however, are severely outdated and frankly, irrelevant. One can't assess creativity or creative thought with so-called "paper and pencil tests" (an elision)—regardless of whether the test material is verbal, nonverbal or figural—and whether in a human or in a human-made machine.

Indeed, overall, the TTCT lacks any overarching theory of creativity or creative thought and has absolutely no relation to real-world creativity (e.g., Newton, Einstein, Picasso, Bach, Graham, Dickinson, Ellison, Morrison, Kafka or anyone or anything else).

Longitudinal studies of the TTCT at 22, 40, and 50 years, moreover, show little relation to any form of "public achievement," that is, creativity recognized by the larger community (Runco et al., 2010).

So, the finding actually tells us virtually nothing—either about artificial "intelligence" or human intelligence—in the realm of creative thought.

### A10a. So, What is Creative Thought?

Let's step back and look at the larger intellective system that makes up the human mind/brain.

It has been argued that there are two major "systems of thought" in humans: a (1) pattern-based extraction system known as "pareidolia" and a (2) rule-based system—combinatorial/componential, hierarchical, and recursive—informed by rules or "algorithms" (Seitz, 2019).

In linguistics, recursion is defined as the ability to embed a noun phrase within another noun phrase or a clause within a clause. But it is this ability to think within a nested hierarchy of related concepts or ideas that is argued to be a unique feature of human thought. This rule-based system is therefore structured into different levels—hierarchical—and it is combinatorial. That is, individual elements may be arranged in a potentially infinite range of combinations and each combination of elements has a different meaning recoverable from the rules or algorithms that inform its composition (Berwick & Chomsky, 2017).

*And recursion appears to be unique to biological systems*. Irene Pepperberg (1999) has noted that parrots (*Psittacines*) can form a new word such as "banerry" by combining the words "banana" and "cherry." But it also looks like they possess the capacity for recursion. It is informally defined as "a process a procedure goes through when one of the steps of the procedure involves invoking the procedure itself." When asked, "What color is the item that is circular and rawhide?" parrots are able to generate a logical conjunction to connect features of objects (circular and rawhide) after dividing the question into its parts (color, shape, texture) and repeatedly using each part to formulate their answer.

Similarly, in natural language processing a sentence can be structured so that whatever follows the verb is just another sentence. In the sentence,

"Barbara believes that rabbits are smart," the first noun and verb form a sentence with an object, "Barbara believes ______," that is followed by another sentence, "Rabbits are smart," that is an object of the first. So, the second sentence is embedded within the first and is formed recursively. This is considered to be a fundamental property of language, mathematics, music, and other symbol and action systems including complex movement (Seitz, 2019). However, LLMs do not appear to have any "understanding" of logical rules or the capacity for recursive thought.

"Based on our analysis, it is found that LLMs do not truly understand logical rules; rather, in-context learning has simply enhanced the likelihood of these models arriving at the correct answers. If one alters certain words in the context text or changes the concepts of logical terms, the outputs of LLMs can be significantly disrupted, leading to counter-intuitive responses" (Yan et al., 2024).

So, you might be curious to ask: Is my pet parrot smarter than an LLM? https://en.wikipedia.org/wiki/Parrot.

On the other hand, Daniel Kahneman, a "behavioral economist," distinguishes between something that he calls System 1 thinking, which is intuitive and immediate, from System 2 thinking, which is deliberate and logical (Kahneman, 2013). However, his model doesn't capture the pattern recognition kinds of thought that organic beings are particularly good at—"image detection" in digital computers—as well as the myriad types of thought processes employed by humans and other animals.

What Professor Kahneman is calling an intuitive and immediate, that is, an elemental cognitive and affective process, may actually be an instinctual response or an *ancestral memory of a highly plastic learned response to the environment* that was sculpted into an instinct by way of evolution. Intuition doesn't tell us anything useful because intuition itself is poorly understood and might actually be quite different from how it is generally defined as "reaching conclusions based on nonconscious processes of reasoning," which tells us very little. But, plasticity-first models may help explain the expeditious evolution of behavioral and anatomical changes in higher-order primates and other animals (Liscovitch et al. 2017).

Moreover, just because something is immediate doesn't mean it isn't deliberative because we often don't know, understand or recognize in ourselves or others the history of thinking behind a topic or thing that we have been thinking about, or about to think about, as the end stage of some reasoning process. C. S. Pierce, the American logician, maintained that *intuition* was actually a mirage because all mental action derives from inference and there is no cognitive stage that precedes all others. Since thinking is always thinking in signs (i.e., symbols) according to Pierce, a thought can only be interpreted in what came before it and in what came after it or otherwise it is bereft of any semantic content or meaning (Pierce, 1877). Indeed, Pierce even had a different take on "introspection" as it is not merely an internal soliloquy with ourselves, but an inference from external, objective knowledge or what he called, "abductive inference" (Pierce, 1976/1914).

Second, these two systems—pattern-based and rule-based—are spread across multiple "noetic forms" or groupings of complex cognitive abilities that include language, number, visuospatial understanding and "image-making," music, instrumental or physical intelligence, emotional expression and social understanding, as well as the ability to classify natural forms found in nature such as animal and plant species (Gardner, 1993).

Let's step back and look at the evolutionary precursors of these abilities.

When Animalia first evolved in the sea and on land, a "sensory manifold" emerged with multiple sensory receptors. A second dual circuit then emerged to segregate internal sensation from this sensorium. Sequencing, organization, and memory of the past thus became "embodied" and evolved to reflect similar qualities of the external world, a central feature of biological systems.

"Supramodular" systems followed. That is, "cross-modal perception" arose and connected these independent sensory systems. Sensory systems do not operate completely independently, but rather sensory information is commingled in the brain, a process referred to as cross-modal perception or "synesthesia." This supramodular system has four defining features (Seitz, 2019).

a. The ability to relate sensory qualities across different sensory modalities.
b. The ability to link an inanimate object to an emotion.
c. The ability to associate a sensory quality to an abstract property.
d. The ability to transform or relate one movement to another.

*Conceptual primitives* appeared and involved an incipient understanding of number (subitization and counting), objects ("naive" or folk physics), geometric forms, places, and navigation (geometry of the environment), instrumental actions of agents, and commonsense or "folk psychology" in which social beings are understood as engaging in actions to reach social goals, form and attend to coalitions, and categorize oneself and others into groups (Spelke, 2022).

Essentially, Professor Spelke is asking what are the basic cognitive capacities that human beings the world over are born with, excepting severe pathology. That is, what do human infants know when they first encounter the world before learning begins? And what knowledge is innate in humans and how do they go about learning about their world? And importantly, are these the same core cognitive capacities that exist in all human cultures around the world that empower a newborn infant to acquire a native language and the unique values, skills, and concepts of the culture and society in which s/he was born?

Conceptual primitives represent what young infants know very early in life as well as what they quickly come to learn about objects, places, numbers, geometry, people's actions and social engagements, as well as about mental states given their updated knowledge of the world over the first year of life.

In machine learning, these basic features of an artifactual system are known as *priors* or initial facts about an event in terms of its prior probability distribution. In Bayesian statistics, Bayesian inference updates these priors—using prior knowledge in the form of a prior distribution—with new information in order to obtain the posteriori probability distribution, which is the conditional distribution of an uncertain quantity given new data. But a comparison between the two—conceptual primitives in humans and initial facts about an event in terms of its initial prior distribution in an LLM—is a conceptual stretch and aleatory at best.

Coevally, memory systems began to differentiate into semantic, episodic, affective, social and collective, prospective, numerical, verbal, visuospatial, kinesthetic, and musical mnemonic systems further underpinning these noetic forms.

Prospective memory is event- or time-based and involves remembering to perform, for instance, an activity under specific circumstances or at a particular point in time. Collective memory is the pooled knowledge of a particular social group.

Then, about 350 million years ago, "affective consciousness" appeared in tetrapods (reptiles, birds, mammals) when early Animalia began to consciously experience affective states. Biological rhythms, the bedrock of many bodily and nervous system activities, became linked with sound as they were prerequisites for nonverbal behaviors and actions such as aesthetic movement (dance) and music (Seitz, 2002, 2005b). Organizational and planning abilities emerged enabling animals to anticipate and plan the future around changing environments and fluctuating climates.

The emergence of full-blown *embodied thought* followed: Manual dexterity, as well as tool use, rapidly evolved. Oral and vocal dexterity laid the foundation for complex syntactical speech. Similarly, self-awareness (i.e., secondary consciousness) evolved in a subset of mammals, primates and potentially some avians, along with the capacity for cognitive and behavioral self-monitoring enabling robust working memory, metacognition (reflection on one's own thoughts), and metalinguistic awareness (reflection on one's own words).

Over millions of years, "paradigmatic thought" or categorization of the natural world arose first in the ability to detect patterns (pareidolia) and was one of the essential ways that organisms gained information and acquired knowledge about physical reality and the natural world. Much later, it became the core of image-making or visuospatial thought that goes back in modern human history to, at least, 120 thousand years ago based on recent discoveries of human artifacts in sub-Saharan Africa (Seitz, 2019).

But it first arose in the ability to extract statistical regularities from the environment, to extract and parse information into meaningful wholes, as well as to predict and generalize from a subset of external stimuli relying on similarity and temporal and spatial contiguity. This latter associative system would be useful in unknown and constantly changing environments where it could use distorted, degraded, and incomplete information to extract these regularities.

Precursors to symbolic thought included "indices," or some physical connection to an object — such as animal tracks in the snow — and "icons," or some likeness or semblance to an object — such as a human face in the clouds.

"Syntagmatic or narrative thought" or the conveyance and telling of stories, arose first in gesture and then was mapped onto sounds and later became reified in the complex architecture of speech. Creative thought emerged on the back of the supramodular systems. This was the underpinning of *metaphoric thought* or *the ability to think of one thing in terms of another* or what has been referred to as the *kernel of creative thought* (Seitz, 1998, 1999, 2000, 2002, 2005a, 2005b, 2019, 2023).

But extant forms of artificial "intelligence" have none of these abilities. And "creative" output by an LLM or other machine "learning" systems is merely stochastic.

Nevertheless, the default and salience networks, the empathy and amygdala circuits, the underlying cerebral architecture for theory of mind, as well as further embodiment of thought itself evolved (see below under (2)). Sociality exploded including such key features as social hierarchies, dominance relations, social roles and relations, social understanding, and Machiavellian intelligence, particularly in early hominid cultures.

A full rule-based system followed involving manipulations on relations among symbols. At a minimum, this rule-based system included general algorithms or rules; cognitive strategies or "heuristics;" deductive, inductive, and abductive reasoning; causal reasoning including protoconditional inferences (if p, then q), protonegation (if not p, then not q), and the incipient ability to comprehend disjunctive syllogisms (if either p or q is true, and p is false, then q is true); conditional and pragmatic (practical) reasoning; categorical and conceptual inference; decision-making; reasoning by analogy; schemas and scripts for obtaining information about the world and using that information to influence that world; local and global planning; insight and creative thought; wisdom; as well as serial understanding to achieve a pre-defined end or praxis (Seitz, 2019).

### A10b. So, What is "Intelligence?"

One of the fundamental issues in the biological and social sciences is that the concept of intelligence is not well understood, and it is not clear what intelligence actually is in humans, in other animal species or if it even exists in any artifactual system such as a computing machine.

Here are some suggestions:

"To my mind, a human intellectual competence must entail a set of skills of problem solving — enabling the individual to resolve genuine problems or difficulties that he or she encounters and, when appropriate, to create an effective product — and must also entail the potential for finding or creating problems — and thereby laying the groundwork for the acquisition of new knowledge" (Gardner, 2022).

"A very general mental capability that, among other things, involves the ability to reason, plan, solve problems, think abstractly, comprehend complex ideas, learn quickly and learn from experience . . . it reflects a broader and deeper capability for comprehending our surroundings— "catching on," "making sense" of things, or "figuring out" what to do" (Stern, 2017).

"Individuals differ from one another in their ability to understand complex ideas, to adapt effectively to the environment, to learn from experience, to engage in various forms of reasoning, to overcome obstacles by taking thought. Although these individual differences can be substantial, they are never entirely consistent: a given person's intellectual performance will vary on different occasions, in different domains, as judged by different criteria. Concepts of "intelligence" are attempts to clarify and organize this complex set of phenomena. Although considerable clarity has been achieved in some areas, no such conceptualization has yet answered all the important questions, and none commands universal assent. Indeed, when two dozen prominent theorists were recently asked to define intelligence, they gave two dozen, somewhat different, definitions" (Neisser et al., 1996).

So, without a clear consensus about what intelligence actually is, how can we clearly delineate its role in humans, non-human species or even in a computing machine?

## A11. Human Brains Versus Computer Chips

The modern human adult brain contains, on average, about 86 billion neurons of which approximately 16.34 billion (19%) make up the grey and white matter of the cortex. The current estimates are there are about 6 billion+ neurons in the gray matter of the cortex and 10.3 billion neurons in the white matter with supporting neuroglia in both (8.7 billion and 20 billion, respectively) (Seitz, 2019).

Moreover, the existing estimates are that there are about 100 trillion synaptic (cell-to-cell) connections in the human adult brain. It's not quite, but almost all about the connections and well over 1100 times more than the number of neurons themselves. If the number of neurons in the typical corvid brain is approximately 1.74 billion—the neurons have an unusual compact structure—then the number of connections would hover around 1.91 trillion. That's one reason why they are smart.

For comparison's sake, the Cerebras CS-2 system—the largest computer chip ever built—8.5" on its side, contains 2.6 trillion transistors, and 850,000 AI-optimized cores is a monstrosity (Cerebras Systems, 2024). But it's the exact opposite of a human or corvid brain: (1) the transistors/cores are nearly all the same in the CS-2 system while human and corvid brains have dozens of different types of specialized neurons that perform diverse functions, (2) the number of connections between cores in the CS-2 system is a small fraction of number of connections in a human or corvid brain—there are up to $10^4$ connections in any one neuron to both adjacent and distant neurons—and (3) these human and corvid connections are further modulated by unique neurochemicals and supporting structures and processes. That's three good reasons why computers aren't "intelligent."

And how might any existing artificial "intelligence" amplify human intelligence?

### A12. Augmenting Human Intelligence

Michael I. Jordan, Department of Computer Science, University of California, Berkeley, posits "intelligence augmentation" as the current state of "human-imitative AI." Computers and software programs are useful because they augment our own intelligence just as a simple hand calculator or digital clock still does (Jordan, 2019).

Although it has been referred to as "artificial intelligence" (AI), it might be better thought of as "synthetic intelligence." A synthesis of the presumptive ways that the human brain and human cognition work together to create human-like "intelligence."

Indeed, this intelligence augmentation involves an intelligent infrastructure made up of data, computation ("software"), and physical entities ("hardware") that "make human environments more supportive, interesting and safe."

## A13. Conclusions

But, looking ahead, why would we want "artificial intelligence" to just mimic human intellect? Augmenting biological intelligence? Great. But simply emulating human intelligence would have limited advantages.

What we want to create is a new kind of artificial "intelligence," a "synthetic intellect," that will abet human creativity and intelligence.

Not a mental doppelganger, but a new class of factitious mentality, that thinks and creates in unfamiliar and novel ways.

Syncretism not simulacrum.

Indeed, a distinguished professor of artificial intelligence expresses similar sentiments:

"The concept of "AGI" (a system that can match or exceed human performance across all tasks) shares all of the defects of the Turing Test. It defines "intelligence" entirely in terms of human performance. It says that the most important AI system capabilities to create are exactly those things that people can do well. But is this what we want? Is this what we need? I think we should be building systems that complement people;

systems that do well the things that people do poorly; systems that make individuals and organizations more effective and more humane. Examples include: 1. writing and checking formal proofs (in mathematics and for software), 2. writing good tests for verifying engineered systems, 3. integrating the entire scientific literature to identify inconsistencies and opportunities, 4. speeding up physical simulations such as molecular dynamics and numerical weather models, 5. maintaining situational awareness of complex organizations and systems, 6. helping journalists discover, assess, and integrate multiple information sources, and many more. Each of these capabilities exceeds human performance -- and that is exactly the point. People are not good at these tasks, and this is why we need computational help. Building AGI is a diversion from building these capabilities." – Thomas G. Dietterich (Distinguish Professor, Emeritus, Oregon State University; Former President, Association for the Advancement of Artificial Intelligence, 04.16.2024, "tweets" on Twitter (X), https://engineering.oregonstate.edu/people/thomas-g-dietterich).

## 3. Are Computing Machines Sentient or Conscious?

## B. What is Consciousness and Sentience?

For a related set of reasons, computers and their software programs are not sentient or conscious.

We know that perceptual consciousness—primary consciousness or awareness of one's environment—may have arisen in vertebrates as early as 530 million years ago (Feinberg & Mallet, 2017). But self-awareness or secondary consciousness may have evolved only in complex vertebrates including canines; cetaceans such as dolphins, porpoises, and whales; chimps and bonobos; elephants; and possibly corvids such as magpies (Seitz, 2019).

Yet modern humans also have the capacity to think about thinking or engage in complex *metacognition* and that appears to be unique to the animate but not inanimate worlds (e.g., digital computers). Indeed, secondary consciousness or self-awareness would have enabled *inner thought or consciousness of one's own thought processes* with emergence of the brain's default and salience networks. These networks included the mind-wandering network ("default"), as well as the network for filtering and detecting relevant stimuli ("salience"). Those networks would have created the capacity for introspection—the ability to examine one's own conscious thoughts and feelings, inferring the actions and intentions of others, as well as myriad nonconscious processes underlying complex cognitive and affective faculties (Dehaene, 2014).

Nonetheless, while *sentience* is the capacity to experience sensations and emotions, "an organism [or computing machine] has *conscious mental states*, [if and only if], there is something that it is like to *be* that organism – something that it is like *for* the organism" (Nagel, 1974).

And only *that* particular organism or thing can have *those* unique conscious states. But there is no evidence that "computing machines" of any kind, think or reason, thus lacking any sort of conscious mental states. Or, to put it another way, if they are not conscious, what does it mean to say that they think or reason?

Antonio Damasio (1996, 1999) suggests that bodily signals—visceral sensations emanating from the abdominal area as well as bodily sensations arising from other parts of the body—modify the way the brain handles information, and this is accomplished primarily through our prefrontal cortices in the thinking part of our brains. Because they lack these bodily signals or *somatic markers*, patients with damage to the frontal lobes are unable to even anticipate the future. That is, they are unable to predict future events, such as make simple, ordinary choices in everyday life, because they possess no biasing somatic state to rely on in decision-making.

These proprioceptive, sensory, and autonomic signals are incorporated in the creation of consciousness as Damasio (1999) as well as Edelman and Tononi (2000) have separately proposed. In both Edelman & Tononi's and Damasio's formulations, nonverbal preconscious and conscious states arise from the mapping of bodily states by numerous brain systems. These include brain stem nuclei, sensory cortices, hypothalamic and basal forebrain areas, vagal nerves originating in the abdominal areas, as well as the parietal cortex where the brain forms bodily maps of the musculoskeletal system with feedback from our internal milieu and musculature. Cingulate cortices and other midbrain areas—especially the superior colliculi, which collates multisensory information—provide mental representations of the relation of one's bodily self to the environment.

Higher-order conscious states such as self-awareness arise from these preconscious and conscious representations of bodily states and their relation to the environment. *Thus, according to these researchers, consciousness is bodily-based and thus would not be amenable to subsumption in silicon or some other non-body-based system.* Indeed, Edelman and Tononi argue that higher-order conscious states are created by the integration of the present with the remembered past. The present consisting of perceptual categorization of incoming sensory stimuli.

### B1. Consciousness in Canines

Canines and some other animal species manifest primary consciousness but, in some limited cases, potentially primary and secondary consciousness.

There is now some preliminary evidence from the Dog Cognition Lab at Barnard College/Columbia University that certain breeds of canines—or even certain individual dogs—display self-recognitory abilities or something like "self-awareness" (Horowitz, 2017). That is, they clearly differentiate themselves from other canines in an *"olfactory mirror test"* (the olfactory version of the *mirror recognition test*, see below).

So, from recent research studies on canines they appear to think (e.g., dogs often tilt their heads before they retrieve a named toy suggesting that it might signify concentration and recall, Sommese et. al., 2021), they appear to be sentient—expressing pain to an injured body part, "howling" for a lost companion—and they may evince "self-awareness" as evidenced in an olfactory mirror test, as well as have something like a "theory of mind" (I wonder if Tom wants to walk to the park?).

> "In particular, if the animal seems to be operating with regard to some mediating element between others' appearance and their behaviors, this behavior could be described as a rudimentary theory of mind" (Horowitz, 2011).

They also understand referential gestures such as pointing—known as a deictic gesture—unlike chimps and bonobos in the wild. But the latter may have arisen as the result of domestication and the myriad ways that humans and other species learn from one another.

## B2. Consciousness in Avians

In birds and mammals, pallial neurons are functionally organized similarly, with sensory and effector areas, richly interconnected hubs, and highly associative areas in the hippocampus and "nidopallium caudolaterale"—equivalent to the monkey prefrontal cortex—which is the portion of the pallium that is the seat of our (and their) ability to act on thoughts, feelings, and make decisions and informed by current reality gained from the senses (Seitz, 2019).

And fibers within and across bird pallial areas are mostly organized at right angles reminiscent of the orthogonal, tangential, and radial organization of mammalian cortical fibers.

We know that corvids (like crows) have sensory neurons that represent numeric quantities. And the associative pallium of crows, like the macaque prefrontal cortex (an old world monkey), is rich in neurons that represent what the animals next report to have seen—whether or not that is what they were shown (Nieder, 2017).

Thus, it appears that birds have (sensory) consciousness or patterns of neuronal activity that represent mental content that drives behavior. Nonetheless, there is some limited evidence that magpies—another corvid species—have demonstrated self-awareness, that is, the ability to recognize themselves in a *mirror recognition test*. In this case, they will try to remove a mark placed on their beak that they notice in a mirror in their cage. But the neuroarchitecture, particularly the connections between structures, explains why birds are as cognitively talented as mammals.

## B3. Consciousness in the Octopus

One may think of consciousness as an active relationship with the natural world with distinct faculties for perceiving, acting, and remembering past events. Indeed, it appears that octopuses have sentient qualities in their ability to protect a body part that has become injured— and presumably conveys pain—suggesting that they are sentient to some extent (Godfrey-Smith, 2016).

But they also demonstrate play, tool use, behavioral flexibility, curiosity, and deception. They possess a distributed nervous system (essential parts of the nervous system are in the eight arms), engage in social learning, have recognition memory for other octopuses, and display episodic memory—memory for events such as food availability. Moreover, their "deimatic" displays on their skin (i.e., "startle" or aposematic displays used to warn, repel or mimic) may even encompass a "visual language" that they use to communicate with others in the oceans.

So, it looks like cephalopods like octopuses are sentient and—given their extensive cognitive abilities—may possess something like primary (sensory) consciousness and maybe even vestiges of self-awareness.

## B4. Consciousness in Human Infants and Young Children

When do human infants become self-aware? Phillipe Rochat at Emory University has suggested five levels of self-awareness that unfold from birth to five-years-of-age (Rochat, 2003). If you place a human infant in front of a mirror before the second year, they display characteristic responses such as emitting a social smile, exploring the image in the mirror and its movements often with delight, and cooing. But by two-years-of-age, toddlers display a completely different set of responses including completely halting their body movements out of intense interest in the object in the mirror, and often hide their faces or tuck their heads in their shoulders, displaying characteristic embarrassment. This is the classic "mirror recognition test" as well as the "mirror mark test" where a mark is placed on the infant (or a primate such as a chimpanzee) to see whether they go on to explore it in the mirror. Self-awareness and self-consciousness are on vivid display. And by three-to-five years of age they go on to develop a robust "theory of mind" about others (see (C) below).

## B5. Conclusions

So, there is a broad array of scientific evidence that sentience and consciousness (both sensory consciousness—awareness of the environment—and self-awareness) are widely distributed across the animal kingdom including primates, mammals (e.g., canines), avians (sentience; primary and possibly secondary consciousness), and cephalopods (e.g., octopuses). And numerous bodily states may play a causal role in producing it.

Any potentially sentient or conscious system needs to be able to weave a huge trove of information from the bodily and nervous systems—whether in a human or a honeybee, of which the latter may possess some aspects of sentience due to evolutionarily ancient midbrain structures shared across species (Barron & Klein, 2016)—into a coherent conscious representation of the world.

But there is not a shred of scientific evidence that LLMs or other artificial "intelligence" systems or "computing machines" possess or display sentience or consciousness.

So, you might be hesitant to ask: What does it feel like to be a Roomba (http://bit.ly/4bymjpu)?

## 4. Do Computing Machines have a "Theory of Mind?"

## C. What is "Theory of Mind?"

The capacity to understand what other individuals may be thinking or feeling is known as "theory of mind" (ToM). That is, cognizing another's putative intentions, beliefs, desires, and even what they may be currently thinking about (Baron-Cohen, Leslie, & Frith, 1985, Premack & Woodruff, 1978; Wimmer & Perner, 1983).

It is called a "theory" because "a system of inferences of this kind is properly viewed (as such) because such states are not directly observable, and the system can be used to make predictions about the behavior of others" (Premack & Woodruff, 1978).

Theory of Mind (ToM) has numerous underlying mechanisms that contribute to understanding what others may be thinking, feeling or planning. These include noticing social cues such as facial emotion and vocal prosody as well as interpreting a character's intention in a social scenario; shared knowledge about the world such as the ability to attribute mental states and mentalistic explanations to literary characters; and facility in interpreting actions such as behavior based on false beliefs, appearance-reality distinctions, and other kinds of misunderstandings, to name a few (Byom & Mutlu, 2013).

### C1: False Belief Tasks

For instance, in a first-order false belief task:

"A child is shown a scene with two doll protagonists, Sally and Anne, with a basket and a box, respectively. Sally first places a marble into her basket. Then Sally leaves the scene, and in her absence, Anne moves the marble and puts it in her box. Then Sally returns, and the child is asked: "Where will Sally look for her marble?"

In the first-order false belief task, the child is simply asked about Sally's belief about the location of the marble not what another child may believe.

But, in a second order false belief task:

"A child is shown a scene with two doll protagonists, Sally and Anne, with a basket and a box, respectively. Sally first places a marble into her basket. Then Sally leaves the scene, and in her absence, Anne moves the marble and puts it in her box. However, although Anne does not realize this, Sally is peeking through the keyhole and sees what Anne is doing. Then Sally returns, and the child is asked: "Where does Anne think that [Sally will] look for her marble?"

Thus, this is called a second-order false belief task because it involves nested beliefs (i.e., beliefs about beliefs). What does Anne think about what Sally is thinking?

## C2. When and How do Children Acquire a "Theory of Mind?"

While second-order false belief tasks are easily handled by normal six-year-olds, and first-order belief tasks by normal four-year-olds, autistic children and normal children under three-years-of-age appear to lack it, whereas adult infra-human higher-order primates (e.g., chimpanzees, bonobos) may possess it.

Nevertheless, young children establish the capacity for theory of mind between three and four-years-of-age and are able to engage in deceit

(e.g., my mom won't know that I ate a cookie while she was upstairs) as well as say one thing and mean another (e.g., I don't want to play anymore…because I'm hungry for dinner).

How so? The attention of others directed towards oneself, joint attention between two individuals directed to another individual or thing, and the act of pointing to reference a person or object, can induce beliefs about another's behavior or intentions and stimulate an interest in another person's mental state in young children.

### C3. Do Other Species or Computing Machines Acquire a "Theory of Mind?"

So, if a bonobo sees a human hide an object—the experimenters track the bonobo's gaze—in the presence of another human who then leaves the room and the object is removed, the bonobo understands that the second human will think that the object is still there in what is clearly a *false-belief task*. Similar behaviors are seen in scrub jays (another member of the family of corvids) who re-cache foods to avoid pilfering by other birds.

But if a computing machine, as I have argued, is incapable of thought or reason, as well as lacks any capacity for sentience or consciousness, how could it even begin to understand what others are feeling or thinking, their intended actions, that is, possess a theory of mind? In fact, the question itself is completely counterfactual.

### C1. Conclusions

None of this—first- or second-order false belief tasks— or theory of mind more generally, is possible in computing machines because they are (1) incapable of thought or reason and (2) lack any capacity for sentience or consciousness.